# Networkwide Traffic State Forecasting Using Exogenous Information: A Multi-Dimensional Graph Attention-Based Approach

Syed Islam
Department of Civil and Environmental Engineering
University of Connecticut, Storrs, CT 06269
Email: syed.islam@uconn.edu

Monika Filipovska
Department of Civil and Environmental Engineering
University of Connecticut, Storrs, CT 06269
Email: monika.filipovska@uconn.edu

**ABSTRACT**
Traffic state forecasting is crucial for traffic management and control strategies, as well as user- and system-level decision making in the transportation network. While traffic forecasting has been approached with a variety of techniques over the last couple of decades, most approaches simply rely on endogenous traffic variables for state prediction, despite the evidence that exogenous factors can significantly impact traffic conditions. This paper proposes a multi-dimensional spatio-temporal graph attention-based traffic prediction approach (M-STGAT), which predicts traffic based on past observations of speed, along with lane closure events, temperature, and visibility across the transportation network. The approach is based on a graph attention network architecture, which also learns based on the structure of the transportation network on which these variables are observed. Numerical experiments were performed using traffic speed and lane closure data from the California Department of Transportation (Caltrans) Performance Measurement System (PeMS). The corresponding weather data were downloaded from the National Oceanic and Atmospheric Administration (NOOA) Automated Surface Observing Systems (ASOS). For comparison, the numerical experiments implement three alternative models which do not allow for multi-dimensional input. The M-STGAT is shown to outperform the three alternative models, when performing tests using our primary data set for prediction with a 30-, 45-, and 60-minute prediction horizon, in terms of three error measures: Mean Absolute Error (MAE), Root Mean Square Error (RMSE) and Mean Absolute Percentage Error (MAPE). However, the model's transferability can vary for different transfer data sets and this aspect may require further investigation.

**INTRODUCTION**

Traffic state reveals important information about the traffic conditions and the performance of a particular road segment or network. Knowledge of the traffic state is fundamental for implementing traffic control and operation strategies, as well as informed user- and system-level decision making (*1–3*). In the era of Intelligent Transportation Systems (ITS), it has become essential to predict the traffic state with a higher level of accuracy and reliability (*4*) to enable better decision making for the efficient control and management of traffic.

Traffic state can be estimated from a variety of traffic variables such as speed, flow, density, or other equivalent variables (*2*).Among these, speed is comparatively more convenient for its ease of conversion to other measures of traffic state (*5*), and is directly related to travel time which is often the most important measure for travelers (*5*, *6*). In addition to the traffic variables, traffic state is closely related to various exogenous factors such as special events, road conditions, and weather (*7*). Specifically, lane closure is one exogenous factor that can affect traffic conditions by reducing mobility and forming long queues (*8*). Traffic disruptions caused by lane closures can vary by location, type of road, and depend on the number of impacted lanes (*8*). Additionally, traffic state can be influenced by weather conditions (*9*) and their inclusion in traffic state forecasting may improve prediction accuracy (*10*).

Over the last three decades, researchers have proposed various approaches for understanding traffic and predicting traffic state (*11*). Before the advent of big data and advanced computational capacity, statistical approaches were common for traffic state forecasting. Numerous studies applied approaches such as ARIMA and Kalman filtering for predicting macroscopic traffic variables – speed, density, and flow for future time horizons (*12–16*). While the early approaches were limited to predicting traffic state for a single location, subsequent studies introduce approaches for predicting traffic state at multiple locations with space-time ARIMA (STARIMA) and adaptive Kalman filtering (*17*, *18*). However, these approaches can be limited in their ability to solve complex spatio-temporal forecasting problems (*19*), and recent studies have found promising results from deep learning approaches in traffic state forecasting (*20*), which are also effective in capturing the spatio-temporal relationship.

Early machine learning (ML) approaches for traffic forecasting focused on short-term predictions using multi-layer perceptron (*21*), deep belief networks (*22*, *23*), stacking convolutional neural networks (CNNs) and long short-term memory (LSTM) networks (*24*), and evolved to using meta-learner frameworks (*25*). Later, machine learning approaches were used for longer-term predictions, with notable techniques such as deep learning stacked auto-encoders (*26*) and LSTM neural networks (*27*). However, these studies focus on the temporal aspects of traffic state prediction and disregard the spatial component but were nevertheless effective for short-term prediction (*28*). Long-term predictions beyond the 30-minute time horizon are often more complicated, since it can be difficult to foresee the evolution of traffic as it fluctuates with a variety of exogenous and endogenous factors that bring uncertainties in the network (*29*). Recent studies are applying diffusion convolutional recurrent neural networks (RNN) (*30*), spatio-temporal recurrent convolutional networks (SRCN) (*31*) and CNNs (*32*) for spatio-temporal feature extraction from traffic data. Graph neural networks (GNNs) show promising results in capturing the spatio-temporal aspects for mid- to long-term traffic forecasting, as seen with spatio-temporal graph convolutional networks (STGCN) and spatial-temporal graph attention networks (STGAT) (*33–35*).

Most studies on traffic forecasting aim to predict the evolution of traffic temporally or spatially, based on historical and spatially distributed traffic information. However, apart from these endogenous features, external events and conditions can also significantly impact traffic conditions. Essien and coauthors applied a Bi-directional Long Short-Term Memory (LSTM) stacked autoencoder (SAE) architecture for traffic state forecasting using lane closure and weather data and their results suggested that models trained without this information suffer predictive accuracy loss (*36*). Two recent studies incorporated rainfall data for forecasting using deep learning architectures (*37*, *38*), and demonstrated the benefit of incorporating weather data in traffic speed forecasting. Specifically, George and coauthors proposed Fuzzy Optimized Long Short-Term Memory (FOLSTM) for capturing the abnormal traffic conditions that affect traffic speed forecasting (*38*). Shabarek et al. adopted deep learning models for traffic speed prediction under adverse weather conditions (*39*). A study by Abadi and coauthors suggests that a major challenge in traffic state forecasting is capturing the effect of exogenous factors such as the occurrence of lane closures, traffic incidents, and special events (*40*). While weather information, besides rainfall, has rarely been incorporated into traffic forecasting, studies have emphasized the importance of considering weather conditions to understand various traffic phenomena outside the context of traffic prediction (*41–45*).

To incorporate exogenous factors in traffic state prediction, with the goal of improving the performance of prediction models and their transferability to previously unseen conditions, this study proposes a multi-dimensional spatio-temporal graph-attention traffic prediction (M-STGAT) approach. The contributions of this study are in the development of this multi-dimensional model architecture that allows for traffic speed forecasting using exogenous variables. The approach specifically relies on utilizing the structure of the underlying transportation network for traffic prediction and learning the relationships between the exogenous and endogenous variables.

**PROBLEM STATEMENT**

Traffic state forecasting problems typically predict future values for aggregate traffic flow variables (speed, flow, or density) based on past observed values of the same variables across space and time. Other exogenous factors such as lane closures and weather conditions can also affect the traffic stream, and thus may be beneficial for accurate prediction of the traffic state. Therefore, this paper considers the problem of predicting the aggregate traffic speed for a future time horizon based on historical aggregated speed data, lane closure events data, and air temperature and visibility information across the transportation network. Though most studies adopt precipitation as the weather variable, our study uses temperature and visibility due to low propensity of rainfall in our selected study area.

To formalize the problem definition, let the road network be represented as a graph $G(N, A)$ where $N$ is the set of nodes and $A$ is the set of links connecting the node. Unlike typical graph representations of transportation networks, we transform the road network into a graph so that each node $i \in N$ corresponds to one of the $|N|$ detector locations and links $(i, j) \in A$ are created for cases where stations are directly connected by adjacent links on the road network. Link $(i, j) \in A$ connects stations $i$ and $j$ if traffic is flowing from $i$ to $j$ directly, without any other intermediate stations.

Time is discretized into short intervals of equal duration $\delta$ (i.e., time steps). For the prediction task, we define a historical time horizon $H$ consisting of $|H|$ time steps and a prediction horizon $T$ consisting of $|T|$ time steps. The observation for each time step $t$ consists of several variables, each represented as a vector of size $|N|$ capturing the information across the $|N|$ stations, of the form: $X_t = \left[x_t^0, \ldots, x_t^{|N|}\right] \in R^{|N|}$ where $x_t^i \in R$ is the observed value at time $t$ and location $i \in N$. Here, we let $X$ be a general representation of any variable that is observed over space and time in this manner. In this study, we specify four variables $X_t^1$ denoting the aggregated traffic speed observation, $X_t^2$ denoting lane closure observations at the locations of each station, and $X_t^3$ and $X_t^4$ denoting air temperature and visibility measured at each station, for each time step $t$. These vectors form a matrix of observations for a given time step $t$ of the form $\mathbb{X}_t = [X_t^1, X_t^2, X_t^3, X_t^4]'$, which can be generalized to problems with a larger set of variables as well. The goal is to predict the speeds over the future time horizon $\hat{X}_{t+1}^1, \hat{X}_{t+2}^1, \ldots, \hat{X}_{t+|T|}^1$ based on the historical observed data $\mathbb{X}_{t-|H|}, \ldots, \mathbb{X}_{t-1}, \mathbb{X}_t$ by maximizing:

$$logP\ (\hat{X}_{t+1}^1, \hat{X}_{t+2}^1, \ldots, \hat{X}_{t+|T|}^1 | \mathbb{X}_{t-|H|}, \ldots, \mathbb{X}_{t-1}, \mathbb{X}_t)$$

where $P(\cdot)$ is an unknown, dynamic, data-generating process. In this representation, predicted values for $X_t^1 \ \forall\ t \in T$ are obtained since we are predicting speed only, as a function of historical values for $\mathbb{X}_t \forall t \in H$ containing observations for all four recorded variables since $\mathbb{X}_t = [X_t^1, X_t^2, X_t^3, X_t^4]'$. The problem statement and model presented below are defined in a general context for spatio-temporal learning with multi-dimensional observations. We introduce the specifics of the multi-dimensional observations when presenting the data used in this study, in later sections, to allow for the problem and model definition to be applicable for multi-dimensional spatio-temporal learning in a more general sense.

## METHODOLOGY

Traffic state prediction, as defined above, is a time-series forecasting problem, predicting the traffic speed in future time intervals based on observations of traffic speed and additional exogenous information observed in past time intervals. However, the traffic data and other exogenous information about the transportation network vary over time and are also defined over the graph of the transportation network $G(N, A)$. Incorporating time-series traffic data into GNN models has proven challenging, and in this study we adopt the vector embedding approach by Zhang and coauthors to represent the spatio-temporal data (*34*). The data observed at each node $i \in N$ are organized based on a fixed-duration historical time horizon $H$, and at each time step $t$ we must store a vector of observations from the previous $|H|$ time steps, $h_t^i = [\mathbb{X}_{t-|H|}, \ldots, \mathbb{X}_{t-1}, \mathbb{X}_t]$, which in our case is a two-dimensional matrix $h_t^i \in \mathbb{R}^{|H| \times 4}$. These matrices are used to construct the network-wide input:

$$H_S^{|N|} = \begin{bmatrix} h_1^1 & \cdots & h_S^1 \\ \vdots & \ddots & \vdots \\ h_1^{|N|} & \cdots & h_S^{|N|} \end{bmatrix}$$

where $H_S^{|N|} \in R^{S \times |N| \times |H| \times 4}$ and S is the temporal sequence length of the entire data set. The data represented in this format feeds into the proposed GNN-based approach.

There are several suitable approaches for temporal feature extraction and learning the temporal dependencies for time-series prediction. According to the literature review, LSTM networks, a type of RNN, are most appropriate for capturing long-term dependencies (*46*). In this study, we specifically consider LSTM with forget gates (*47*), where the forget gates determine which information should be eliminated from a cell state which acts as memory storage persisting through all iterations in the time-series data. In addition, since we are using a graph-based data set, GNN architectures are particularly suitable for spatial feature extraction, and we employ the GAT architecture with a multi-head attention mechanism to learn the relative importance of each node for the prediction at other nodes. In the context of this study, this can be interpreted as learning a weighted matrix which carries information about how much the traffic conditions and exogenous information at one traffic station will impact the traffic state at an adjacent traffic station in the transportation network (*48*). Finally, since the model includes multi-dimensional input, it is important for the model to capture any interactions in the 4-dimensional input data, containing traffic state and exogenous information, and thus we also include a CNN component.

The proposed model architecture for traffic forecasting with exogenous factors is a multi-dimensional spatio-temporal graph-attention approach for traffic prediction (M-STGAT). The proposed approach specifically accommodates the multi-dimensional data into a framework for spatial and temporal feature extraction, utilizing the graph structure of the transportation network. The architecture of M-STGAT is shown in Figure 1, where the multi-dimensional input feeds first into the RNN block consisting of two LSTM layers for temporal feature extraction, which then leads into the GAT block for spatial feature extraction. The output of this component feeds into a CNN block, performing convolution over the multi-dimensional input, which is then passed into a fully connected linear layer to generate the predictions. We use rectified linear unit (ReLU) activation functions between the various layers.

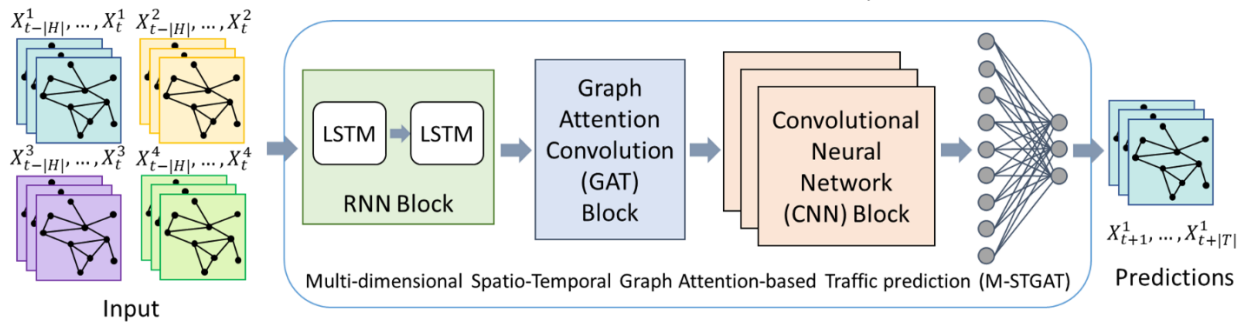

**Figure 1. Model architecture of the proposed M-STGAT**

## NUMERICAL EXPERIMENTS
To test and evaluate the performance of the proposed model architecture for traffic state prediction, we developed a set of numerical experiments using real traffic and exogenous information data on a real-world transportation network.

### Study Sites and Data
The California Department of Transportation (Caltrans) Performance Measurement System (PeMS) collects real-time data in the state of California (*49*) and maintains an Archived Data User Service (ADUS) containing over 10 years of historical data. From the Caltrans data, we use a subset of historical traffic station observations and lane-closure records for district 3 of the 12 districts that they define for the state of California. District 3 is the North Central district in California, covering Marysville in the Sacramento valley, and was selected due to its reported high percentage of functional detectors, according to Caltrans. We supplement the traffic state and lane-closure data with weather variables, specifically temperature and visibility data from the Automated Surface/Weather Observing System (ASOS) stations (*50*). We found 17 weather stations that are located within the Caltrans-defined district 3 and obtained their corresponding weather data.

*Traffic Speed Data Characteristics and Pre-processing*
From the traffic state data available via PeMS, we use 5-minute station data which contains 5-minute aggregated values for speed, flow, and occupancy for each traffic station in district 3, among other information. For the purpose of this study, only the speed data were relevant. In the pre-processing stage, we found that large periods of data, covering most of March and April of 2022, had missing values for most traffic stations, and were thus excluded from the analysis. The data from January, February, May, and June of 2022 are used to create the primary data set for these numerical experiments. Furthermore, in data pre-processing stage we found that only 935 of the 1,454 stations in district 3 had sufficient amounts of data across all four months of the primary data set. Therefore, our data set is based on the information from these 935 stations, whose locations are shown in Figure 2. In the experimental set up, we define three additional data sets for testing the proposed approach, which will be further described in the experimental design.

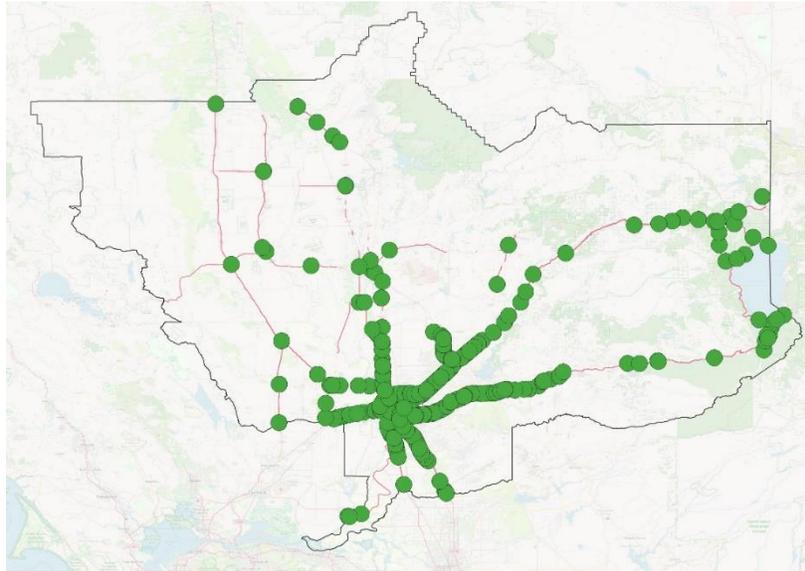

**Figure 2. Locations of 935 traffic stations in district 3, on the corresponding road network**

*Lane Closure Data Characteristics and Pre-processing*
The Caltrans lane-closure data contain detailed information about each lane closure event across the state, including the corresponding route id, direction, begin and end absolute post miles. This information was used together with the traffic station metadata to identify lane closure occurrences at the stations' locations. The traffic station metadata also contains the route id, direction, and absolute post mile to indicate the location of each station and were joined with the lane closure data based on that information. A lane closure event is recorded at a given station if there is a match between the route id and directions in both datasets and if the exact post mile of the station falls between the start and end postmiles of a recorded lane closure event. Temporally, we create a 5-minute lane closure data set to match the 5-minute traffic speed data, where a lane closure event is recorded for all 5-minute intervals that overlap with the time interval for the lane closure event. We define a lane closure variable $l_t^i \forall\ i \in N, t \in S$ as follows:

$$l_t^i = \begin{cases} 0 & \text{if no lane closure is observed} \\ 1 + \dfrac{c_t^i}{n^i} & \text{if a lane closure event is observed} \end{cases}$$

where $n^i$ is the total number of lanes at location $i$ and $c_t^i$ is the number of closed lanes at location $i$ at time step $t$. With this definition, all lane closure events are recorded, including ones where there is only shoulder closure (i.e., the number of closed lanes is zero) in which case the lane closure value is equal to 1.

*Weather Data Characteristics and Pre-processing*

From the ASOS weather data, we extracted the temperature and visibility variables. The information in these data were recorded at irregular intervals and were again reformatted to match the 5-minute intervals of speed and lane closure events data. To match the data spatially, weather information for each of the 935 traffic stations was recorded from the closest of the 17 ASOS weather stations, using the proximity toolset under geoprocessing tools of ArcGIS Pro 3.1.0. The results of this spatial match are shown in Figure 3, where the weather stations are represented with squares, and the traffic stations are represented with circles with the color indicating the nearest weather station. For missing values in the weather data, we applied temporal and spatial interpolation. If the dataset had data missing for consecutive 30 minutes or less, we preformed temporal linear interpolation. In cases where a longer sequence of missing data occurred, the missing data points were filled using the observations at the nearest station with available data at those times.

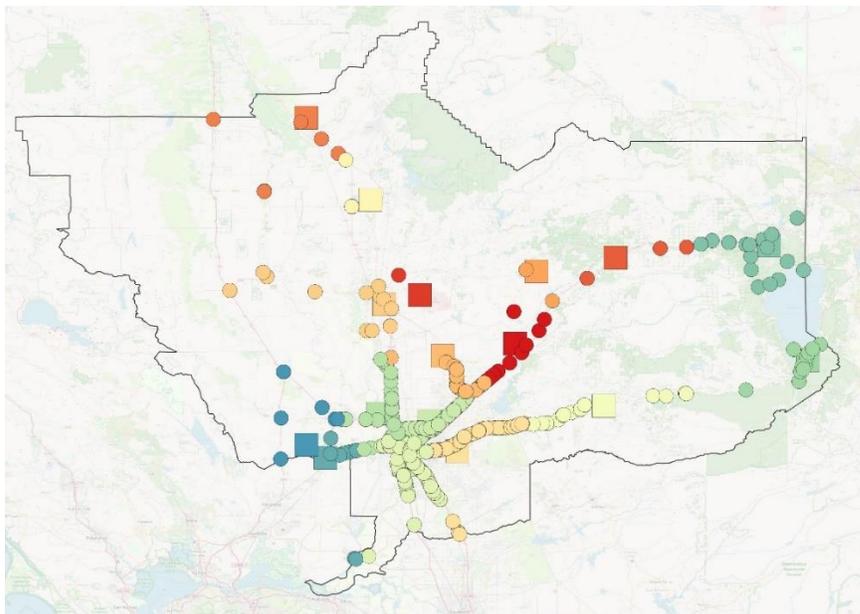

**Figure 3. Locations of the 935 traffic stations and 17 weather stations in district 3**

**Experimental Design**

To evaluate the performance of the proposed approach, we develop a set of numerical experiments which will allow us to answer the following questions:

Islam and Filipovska

1. How does the performance of the proposed multi-dimensional spatio-temporal graph attention-based traffic prediction approach (M-STGAT) compare to simpler one-dimensional approaches?
2. How does the proposed M-STGAT approach perform for various prediction horizons, and how does its performance compare to that of other simpler approaches?
3. How well suited is the proposed M-STGAT approach for zero-shot transfer to be used during different conditions not experienced during training?

To answer the first question, in the numerical experiments we implement three simpler approaches that the M-STGAT can be compared to:

- A one-dimensional spatio-temporal graph attention-based traffic prediction approach (STGAT) which has a similar structure as the M-STGAT but forgoes the multi-dimensional aspect.
- A simple graph-attention network (GAT) model, which also only allows for one-dimensional input and only captures the spatial features for prediction.
- A simple LSTM network with one-dimensional input that only captures temporal features for prediction.

To answer the second question, we perform numerical experiments for models trained and tested with historical and prediction horizons of different durations. Namely, we define three prediction horizons of 30, 45 and 60 minutes in duration, each with a corresponding historical horizon which is twice as long.

Finally, to answer the third question, in addition to testing with the primary dataset (PD) we perform zero-shot transfer tests of the models with three test data sets defined for different time periods and each with different characteristics:

- Transfer data set 1 (TD1): containing data from July and August 2022, where there may be different traffic patterns or exogenous characteristics due to the summer season.
- Transfer data set 2 (TD2): containing data from January and February 2023, where again different weather conditions or traffic patterns could occur but may better match the first two months of the primary data set.
- Transfer data set 3 (TD3): containing data from May and June 2023, where new types of patterns may occur due to another change in seasons.

Further information about the characteristics of the primary data set and the three transfer data sets are shown in Tables 1 and 2. Table 1 presents summary statistics for all continuous variables: speed, temperature, and visibility, including the mean and standard deviation, the minimum, maximum and quartile values observed in the data set. It can be observed that the mean speed values are very close to one another across all four data sets, but they have significant differences in their standard deviations, where the speed standard deviation for TD2 and TD3 are significantly higher than those observed in PD. We found that the mean temperature observed in the primary data set is $58.55°F$, and mean temperatures of $76.10°F$, $46.34°F$ and $64.98°F$ were observed for each of the three transfer datasets, respectively. This variation in weather conditions will allow for an interesting comparison in how the models transfer to the three transfer sets. In terms of visibility, all data sets have a mean close to 9 miles, indicating predominantly clear conditions with high visibility. Nevertheless, TD2 exhibits high standard deviation for visibility, which may be a distinguishing feature of this data set. Table 2 further shows the occurrence of different levels of visibility and lane closure events. We define four levels of visibility, as shown in the table, and can observe that TD2 has a significantly higher occurrence of low visibility and the lowest portion of perfect visibility compared to other data sets. Regarding the occurrence of lane closure events, it is worth noting that though lane closure events are rare in general, TD3 has the highest percentage of lane closure events and was also the data set with the highest standard deviation of observed speeds. Both TD1 and TD3 have a higher percentage of lane closure occurrence relative to the PD set.

Islam and Filipovska**Table 1. Summary statistics of speed, temperature, and visibility for primary and transfer data sets**

| Variable | Data Set | Min | Q1 | Q2 | Q3 | Max | Mean | Standard Deviation |
|---|---|---|---|---|---|---|---|---|
| Speed | PD | 3.00 | 62.70 | 64.90 | 66.90 | 100.00 | 63.89 | 6.03 |
|  | TD1 | 0.00 | 62.60 | 64.90 | 67.20 | 100.00 | 63.45 | 8.67 |
|  | TD2 | 0.00 | 62.40 | 64.90 | 67.30 | 100.00 | 61.63 | 13.84 |
|  | TD3 | 0.00 | 62.10 | 64.90 | 67.20 | 100.00 | 60.38 | 16.18 |
| Temperature | PD | -9.40 | 46.40 | 57.20 | 69.80 | 105.10 | 58.55 | 17.03 |
|  | TD1 | 32.00 | 66.20 | 75.20 | 86.00 | 107.10 | 76.10 | 12.76 |
|  | TD2 | -11.20 | 41.00 | 46.40 | 53.10 | 73.40 | 46.34 | 9.32 |
|  | TD3 | 24.80 | 55.90 | 62.60 | 73.40 | 102.90 | 64.98 | 12.02 |
| Visibility | PD | 0.06 | 10.00 | 10.00 | 10.00 | 10.00 | 9.33 | 2.01 |
|  | TD1 | 0.25 | 10.00 | 10.00 | 10.00 | 10.00 | 9.97 | 0.35 |
|  | TD2 | 0.12 | 10.00 | 10.00 | 10.00 | 10.00 | 9.00 | 2.37 |
|  | TD3 | 0.12 | 10.00 | 10.00 | 10.00 | 10.00 | 9.86 | 0.87 |

**Table 2. Occurrence of different visibility levels and lane closure events for the four data sets**

| Variable | | Dataset | | | |
|---|---|---|---|---|---|
|  |  | PD | TD1 | TD2 | TD3 |
| Visibility Level Occurrences (%) | Low (<7 mi) | 4.64 | 0.07 | 7.07 | 0.68 |
|  | Medium (≥4 and <7 mi) | 3.26 | 0.18 | 5.55 | 0.68 |
|  | High (≥7 and <10 mi) | 6.37 | 1.50 | 7.43 | 3.07 |
|  | Clear (10 mi) | 85.73 | 98.26 | 79.95 | 95.57 |
| Lane Closure Occurrence Events (%) | | 2.57 | 2.86 | 1.44 | 4.45 |

## RESULTS

In performing the numerical experiments, we observed that the M-STGAT and STGAT model required longer training for convergence and thus were trained for 400 epochs, while the remaining models GAT and LSTM were trained for 200 epochs. The prediction accuracy was evaluated using three error measures, mean absolute error (MAE), root mean squared error (RMSE), and mean absolute percentage error (MAPE).

For training, the primary data set (PD) was separated into a training, validation, and test set, containing 72%, 14% and 14% of the data, respectively. Each model used the same training and validation set, and was evaluated using the same test set. We trained three versions of each of the four models, one for each of the 30-, 45- and 60-minute prediction horizons. All results from the PD set are shown in Table 3 (a). From these results, we observe that the M-STGAT model achieves lowest error values across all prediction horizons in training, validation, and testing with the primary data set. Though the STGAT replicates the model architecture of M-STGAT, without the multi-dimensionality, it does not consistently outperform the GAT and LSTM models but seems to have some advantage for longer prediction horizons.

The MAPE values for the four models across are shown in Figures 3, 4 and 5 for the 30-, 45-, and 60-minute prediction horizons, respectively. While no significant pattern can be observed for how most models perform across different prediction horizon durations, and there seem to be significant variations among the three one-dimensional models, it is apparent that the M-STGAT consistently outperforms all three other approaches. This improvement in performance is most significant in the test portion of the primary data set and for the longest 60-minute prediction horizon, where M-STGAT achieves an MAPE

Islam and Filipovska

value of 4.1%, while the next best performing model (GAT) has an MAPE value of 8%. This improvement by 3.9 percentage points is a 48.75% reduction in error relative to GAT.

The transfer testing results are presented in Table 3 (b), showing the values of all three metrics, for the four models and their performance across TD1, TD2, and TD3 for the 30-, 45-, and 60-minute prediction horizons. The MAPE values for transfer testing are visualized in Figures 7, 8, and 9 for the 30-, 45-, and 60-minute prediction horizons, respectively. The relative performance of the four models is not always consistent across the three transfer sets and the three prediction horizons. For TD1, the M-STGAT model can be observed to consistently outperform all other approaches across all three prediction horizons. In this case, the MAPE for M-STGAT is always near 5%, while the second-best approach is the STGAT for the 30- and 60-minute horizons, and GAT for the 45-minute horizon. For TD2, M-STGAT is outperformed by the simple GAT model for all two of the three prediction horizons, where it achieves MAPE lower by 1.13 and 1.64 percentage points for the 30-minute and 60-minute horizons, respectively. Finally, for TD3, the best-performing model for the 30-minute horizon is STGAT for the 30-minute horizon, M-STGAT for the 45-minute horizon, and again STGAT for the 60-minute horizon. For the 45-minute horizon across all three transfer sets, M-STGAT consistently outperforms the three alternative models across all performance metrics.

These results demonstrate the zero-shot transferability of the three alternative models can vary significantly and models that are best performing in some cases can have very high error values in other cases. The M-STGAT approach consistently showed best zero-short transfer performance for the mid-term 45-minute prediction horizon and outperformed the three alternative models in all cases when testing on the test portion of the primary data set. Across all transfer tests, the MAPE values observed for the M-STGAT vary between 5.31% and 12.33%, while those for STGAT span a larger range from 5.9% to 31.27%, for GAT the values are between 7.22% and 11.32%, and for LSTM in the range from 5.08% to 14.85%. On the other hand, in the primary dataset tests, M-STGAT always achieves the lowest errors across all metrics.

1    **Table 3. Error measures from the training, validation, and testing portions of the primary data set and the three transfer data sets**

| Set | Error | (a) Error measures for the training, validation, and testing portions of the primary data set ||||||||||||
|---|---|---|---|---|---|---|---|---|---|---|---|---|---|
| | | 30-min prediction horizon |||| 45-min prediction horizon |||| 60-min prediction horizon ||||
| | | M-STGAT | STGAT | GAT | LSTM | M-STGAT | STGAT | GAT | LSTM | M-STGAT | STGAT | GAT | LSTM |
| Train | MAE | 1.35 | 2.18 | 2.35 | 2.10 | 1.26 | 2.09 | 2.57 | 2.17 | 1.25 | 2.20 | 2.69 | 2.23 |
| | RMSE | 2.27 | 3.63 | 4.21 | 3.61 | 2.10 | 3.45 | 4.57 | 3.78 | 2.08 | 3.64 | 4.78 | 3.91 |
| | MAPE | 2.27 | 5.24 | 4.62 | 3.41 | 2.15 | 5.77 | 9.18 | 4.53 | 2.13 | 3.50 | 6.30 | 4.55 |
| Val. | MAE | 2.37 | 3.94 | 3.06 | 4.16 | 2.20 | 4.02 | 3.38 | 4.27 | 2.21 | 4.22 | 3.65 | 4.34 |
| | RMSE | 4.27 | 7.34 | 5.34 | 7.57 | 4.15 | 7.53 | 5.93 | 7.77 | 4.15 | 7.74 | 6.35 | 7.90 |
| | MAPE | 4.08 | 7.38 | 6.61 | 6.78 | 3.72 | 7.72 | 19.60 | 6.74 | 3.73 | 4.70 | 4.29 | 10.63 |
| Test | MAE | 2.56 | 4.36 | 3.33 | 4.61 | 2.36 | 4.43 | 3.67 | 4.72 | 2.43 | 4.63 | 3.98 | 4.86 |
| | RMSE | 4.59 | 8.08 | 5.78 | 8.25 | 4.45 | 8.19 | 6.40 | 8.49 | 4.52 | 8.45 | 6.90 | 8.77 |
| | MAPE | 4.95 | 8.38 | 5.90 | 8.83 | 3.95 | 8.64 | 4.93 | 4.09 | 4.08 | 10.10 | 7.97 | 9.24 |
| Set | Error | (b) Error measures for testing with the three transfer data sets ||||||||||||
| | | 30-min prediction horizon |||| 45-min prediction horizon |||| 60-min prediction horizon ||||
| | | M-STGAT | STGAT | GAT | LSTM | M-STGAT | STGAT | GAT | LSTM | M-STGAT | STGAT | GAT | LSTM |
| TD1 | MAE | 3.38 | 4.74 | 3.83 | 5.36 | 3.22 | 4.99 | 4.14 | 5.61 | 3.29 | 5.07 | 4.44 | 5.60 |
| | RMSE | 7.46 | 9.73 | 7.47 | 10.35 | 7.39 | 10.04 | 8.00 | 10.62 | 7.39 | 10.10 | 8.43 | 10.71 |
| | MAPE | 5.64 | 6.05 | 7.22 | 13.30 | 5.31 | 9.33 | 7.66 | 11.26 | 5.51 | 5.90 | 8.31 | 13.16 |
| TD2 | MAE | 5.91 | 6.70 | 4.94 | 7.71 | 5.62 | 6.98 | 5.23 | 7.38 | 5.90 | 6.16 | 5.70 | 6.97 |
| | RMSE | 13.47 | 11.65 | 8.93 | 13.07 | 13.24 | 12.17 | 9.36 | 12.74 | 13.33 | 11.33 | 9.79 | 12.55 |
| | MAPE | 9.81 | 31.27 | 8.68 | 13.55 | 9.19 | 15.36 | 9.21 | 12.68 | 9.85 | 11.12 | 8.21 | 5.08 |
| TD3 | MAE | 7.35 | 7.60 | 5.62 | 7.98 | 6.63 | 7.03 | 5.97 | 7.35 | 6.72 | 5.94 | 6.37 | 7.10 |
| | RMSE | 16.08 | 11.99 | 9.04 | 12.89 | 16.02 | 11.83 | 9.49 | 12.43 | 15.96 | 11.25 | 9.90 | 12.56 |
| | MAPE | 12.33 | 8.00 | 10.01 | 14.69 | 10.61 | 14.16 | 10.62 | 14.85 | 10.85 | 9.82 | 11.32 | 12.75 |

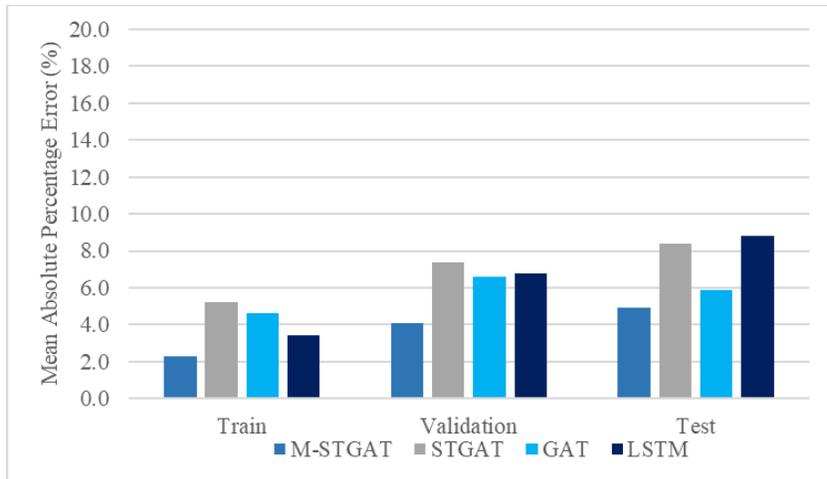

**Figure 4. MAPE values for train, validation, and test partitions of the PD for the 30-minute prediction horizon**

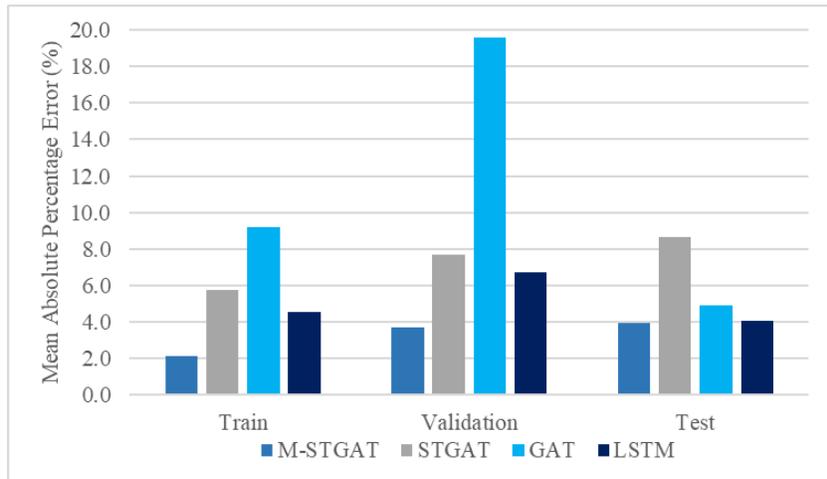

**Figure 5. MAPE values for train, validation, and test partitions of the PD for the 45-minute prediction horizon**

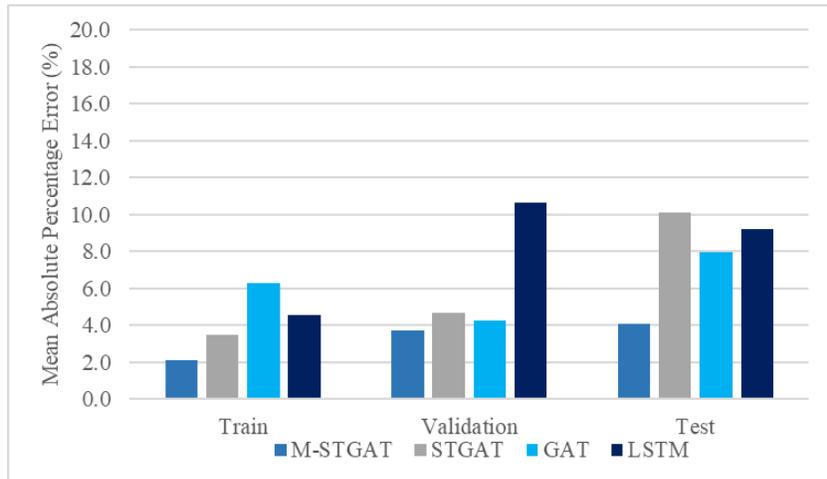

**Figure 6. MAPE values for train, validation, and test partitions of the PD for the 60-minute prediction horizon**

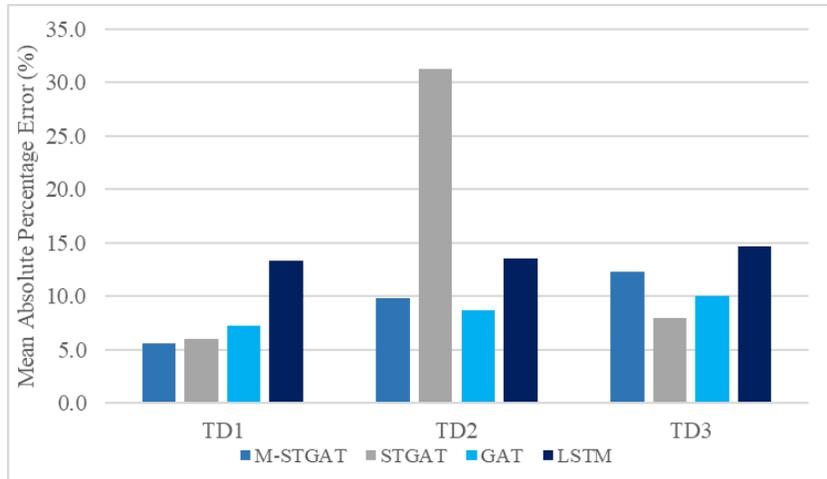

Figure 7. Transfer testing MAPE values for TD1, TD2 and TD3 for the 30-minute prediction horizon

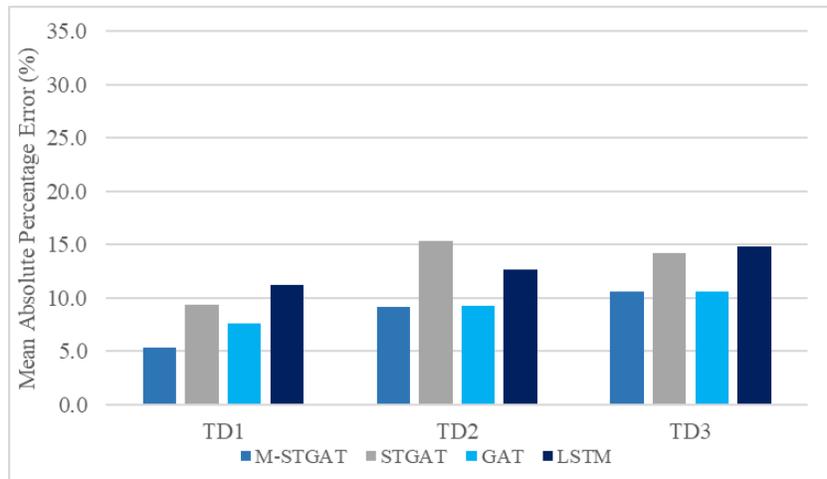

Figure 8. Transfer testing MAPE values for TD1, TD2 and TD3 for the 45-minute prediction horizon

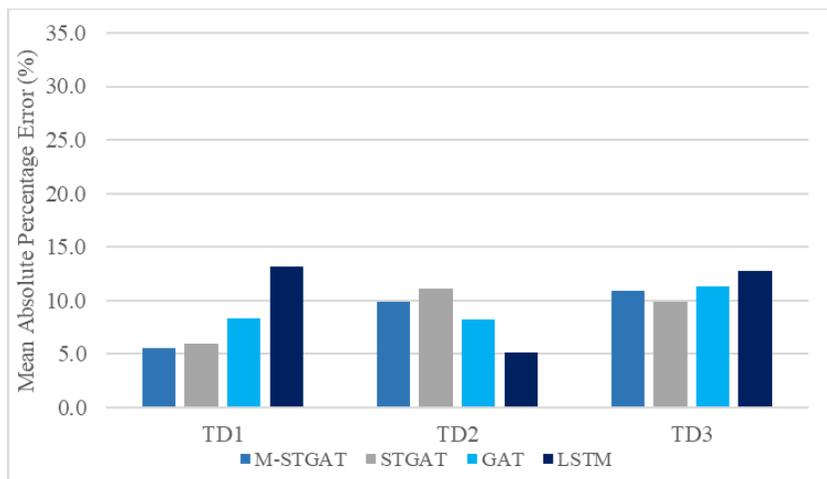

Figure 9. Transfer testing MAPE values for TD1, TD2 and TD3 for the 60-minute prediction horizon

## CONCLUSION AND DISCUSSION

This paper focuses on the problem of traffic state forecasting and presents a machine learning approach that incorporates both endogenous and exogenous information about the transportation network to improve traffic speed prediction. The proposed approach is a multi-dimensional spatio-temporal graph-attention traffic prediction model (M-STGAT), which employs neural networks for spatial and temporal feature extraction, as well as modeling the relationships between the endogenous and exogenous variables. In our implementation, the model's input is four dimensional, containing past speed observations, past lane closure events, and weather information including temperature and visibility throughout the transportation network. The model's structure specifically utilizes these four variables, their relationships with one another and their spatial and temporal interdependencies, along with the transportation network's structure to learn to predict future traffic speeds.

Numerical experiments to evaluate the proposed approach were conducted using the Caltrans PeMS data for traffic state and lane closure information, and ASOS weather data for temperature and visibility data. Specifically, we focused on the north central region (district 3) of California and created 5-minute data across the 935 working traffic stations from speed, lane closure, temperature, and visibility observations. For comparison, in addition to the M-STGAT approach, we implement three alternative models including a one-dimensional STGAT, a simple GAT and an LSTM network. Each of these models are trained using the training portion of our primary data set, and three versions of each model were trained, appropriate for a 30-, 45- and 60-minute prediction horizon, respectively. The models were tested on a portion of the primary data set reserved for testing. In addition to the primary data set, we evaluate the models' transferability using three transfer data sets denoted as TD1, TD2, and TD3.

The results demonstrate that with the primary data set, the M-STGAT approach STGAT consistently outperforms all three other approaches across all metrics. There are no distinct patterns in terms of the models' performance for different prediction horizons, but the most significant improvement in performance in testing with the primary data is for the longest 60-minute prediction horizon where M-STGAT achieves an MAPE value of 4.1%, while the next best performing model (GAT) has an MAPE value of 8%. In transfer testing, the relative performance of the four models is not always consistent across the three transfer sets and the three prediction horizons. For the 45-minute horizon across all three transfer sets, M-STGAT consistently outperforms the three alternative models across all performance metrics. However, in general the results demonstrate that zero-shot transferability of the three alternative models can vary significantly and models that are best performing in some cases can have very high error values in other cases.

One observation from this study, based on the transfer testing results, is that a better-defined training set may be able to improve the models' transferability. In our study, the training set was limited to four months of data, as two of the months initially intended for training had significant amounts of missing data. Furthermore, the results demonstrate that, at least within the primary data set, the impact of exogenous information in the prediction model is significant. While our study did not include precipitation as a weather variable, due to few observed occurrences of precipitation in the study area, precipitation may still be an important variable for traffic state forecasting, especially in regions more prone to precipitation. Future research can also explore additional exogenous variables, such as the cause of lane closure events or other factors that may impact traffic.

## AUTHOR CONTRIBUTIONS

The authors confirm contribution to the paper as follows: study conception and design: S. Islam, M. Filipovska; data collection: S. Islam, M. Filipovska; analysis and interpretation of results: S. Islam, M. Filipovska; draft manuscript preparation: S. Islam, M. Filipovska. All authors reviewed the results and approved the final version of the manuscript.

Islam and Filipovska1736. Essien, A., I. Petrounias, P. Sampaio, and S. Sampaio. A Deep-Learning Model for Urban Traffic Flow Prediction with Traffic Events Mined from Twitter. *World Wide Web*, Vol. 24, No. 4, 2021, pp. 1345–1368. https://doi.org/10.1007/s11280-020-00800-3.
37. Han, D., X. Yang, G. Li, S. Wang, Z. Wang, and J. Zhao. Highway Traffic Speed Prediction in Rainy Environment Based on APSO-GRU. *Journal of Advanced Transportation*, Vol. 2021, 2021, pp. 1–11. https://doi.org/10.1155/2021/4060740.
38. George, S., and A. K. Santra. An Improved Long Short-term Memory Networks with Takagi-Sugeno Fuzzy for Traffic Speed Prediction Considering Abnormal Traffic Situation. *Computational Intelligence*, Vol. 36, No. 3, 2020, pp. 964–993. https://doi.org/10.1111/coin.12291.
39. Shabarek, A., S. Chien, and S. Hadri. Deep Learning Framework for Freeway Speed Prediction in Adverse Weather. *Transportation Research Record: Journal of the Transportation Research Board*, Vol. 2674, No. 10, 2020, pp. 28–41. https://doi.org/10.1177/0361198120947421.
40. Abadi, A., T. Rajabioun, and P. A. Ioannou. Traffic Flow Prediction for Road Transportation Networks With Limited Traffic Data. *IEEE Transactions on Intelligent Transportation Systems*, 2014, pp. 1–10. https://doi.org/10.1109/TITS.2014.2337238.
41. Filipovska, M., H. S. Mahmassani, and A. Mittal. Prediction and Mitigation of Flow Breakdown Occurrence for Weather Affected Networks: Case Study of Chicago, Illinois. *Transportation Research Record: Journal of the Transportation Research Board*, Vol. 2673, No. 11, 2019, pp. 628–639. https://doi.org/10.1177/0361198119851730.
42. Hou, T., H. S. Mahmassani, R. M. Alfelor, J. Kim, and M. Saberi. Calibration of Traffic Flow Models under Adverse Weather and Application in Mesoscopic Network Simulation. *Transportation Research Record: Journal of the Transportation Research Board*, Vol. 2391, No. 1, 2013, pp. 92–104. https://doi.org/10.3141/2391-09.
43. Rakha, H., M. Farzaneh, M. Arafeh, and E. Sterzin. Inclement Weather Impacts on Freeway Traffic Stream Behavior. *Transportation Research Record: Journal of the Transportation Research Board*, Vol. 2071, No. 1, 2008, pp. 8–18. https://doi.org/10.3141/2071-02.
44. Kim, J., H. S. Mahmassani, and J. Dong. Likelihood and Duration of Flow Breakdown. *Transportation Research Record: Journal of the Transportation Research Board*, Vol. 2188, No. 1, 2010, pp. 19–28. https://doi.org/10.3141/2188-03.
45. Mahmassani, H., J. Dong, J. Kim, and R. Chen. *Incorporating Weather Impacts in Traffic Estimation and Prediction Systems*. 2009.
46. Yu, Y., X. Si, C. Hu, and J. Zhang. A Review of Recurrent Neural Networks: LSTM Cells and Network Architectures. *Neural Computation*, Vol. 31, No. 7, 2019, pp. 1235–1270. https://doi.org/10.1162/neco_a_01199.
47. Gers, F. A., J. Schmidhuber, and F. Cummins. Learning to Forget: Continual Prediction with LSTM. *Neural Computation*, Vol. 12, No. 10, 2000, pp. 2451–2471. https://doi.org/10.1162/089976600300015015.
48. Veličković, P., G. Cucurull, A. Casanova, A. Romero, P. Liò, and Y. Bengio. Graph Attention Networks. 2017. https://doi.org/https://doi.org/10.48550/arXiv.1710.10903.
49. California Department of Transportation. Performance Measurement System (PeMS) Data Source.
50. IOWA State University. ASOS Network.